# An Enhanced Prompt-Based LLM Reasoning Scheme via Knowledge Graph-Integrated Collaboration

**Yihao Li, Ru Zhang*, Jianyi Liu**

{yihaoli, zhangru*, liujy}@bupt.edu.cn
School of Cyberspace Security, Beijing University of Posts and Telecommunications

## Abstract

While Large Language Models (LLMs) demonstrate exceptional performance in a multitude of Natural Language Processing (NLP) tasks, they encounter challenges in practical applications, including issues with hallucinations, inadequate knowledge updating, and limited transparency in the reasoning process. To overcome these limitations, this study innovatively proposes a collaborative training-free reasoning scheme involving tight cooperation between Knowledge Graph (KG) and LLMs. This scheme first involves using LLMs to iteratively explore KG, selectively retrieving a task-relevant knowledge subgraph to support reasoning. The LLMs are then guided to further combine inherent implicit knowledge to reason on the subgraph while explicitly elucidating the reasoning process. Through such a cooperative approach, our scheme achieves more reliable knowledge-based reasoning and facilitates the tracing of the reasoning results. Experimental results show that our scheme significantly progressed across multiple datasets, notably achieving over a 10% improvement on the QALD10 dataset compared to the best baseline and the fine-tuned state-of-the-art (SOTA) work. Building on this success, this study hopes to offer a valuable reference for future research in the fusion of KG and LLMs, thereby enhancing LLMs' proficiency in solving complex issues.

## 1 Introduction

In recent years, the rapid development of LLMs (Ouyang et al., 2022; OpenAI, 2023; Thoppilan et al.,2022; Chowdhery et al.,2023; Touvron et al.,2023; Du et al.,2022) has been a significant milestone in artificial intelligence evolution. These models are capable of generating coherent and high-quality responses by pre-training on large text corpora. Moreover, with further supervised fine-tuning, they can be well adapted to various other downstream NLP tasks. However, despite their impressive capabilities, LLMs are still facing several practical application challenges (Kaddour et al.,2023; Koubaa et al.,2023). The first major issue is hallucination. While demonstrating proficiency in generating text, LLMs still struggle with maintaining factual accuracy. They may incorrectly apply their learned knowledge and provide plausible-sounding but wrong outputs (Ji et al.,2023; Brown et al.,2020), which lead to a concerning limitation, especially in critical fields like medical diagnosis and financial decision-making. Secondly, LLMs often face challenges in handling more complex and specialized reasoning tasks that exceed their training scope, while the constant updating of external world knowledge and the substantial resources required for training make it difficult to remain consistently up-to-date. Moreover, due to their complex internal parameters, the opacity of LLMs' reasoning poses challenges to transparency and interpretability (Danilevsky et al.,2020).

Based on the above problems, integrating external knowledge into LLMs, such as Retrieval-Augmented Generation (RAG) via document retrieval or KG retrieval, has become one of the most effective solutions at present. While document retrieval provides relevant information, its effectiveness is hindered by retrieval inaccuracies and document redundancy (Liu et al., 2023b). In contrast, KG offers a rich structured representation of knowledge, and its superior extensibility ensures the continuous update of knowledge (Pan et al., 2023), thus providing LLM reasoning with timely and accurate external knowledge. Recent studies (Li et al.,2023a; Xie et al., 2022; Baek et al., 2023; Yang et al., 2023; Wang et al.,2023; Jiang et al.,2023), which focus on supplementing LLMs with external knowledge from KGs, involve retrieving information to enhance prompt. However, these methods typically relegate LLMs to a passive role, where they primarily receive knowledge without actively participating in the reasoning process, thereby overlooking the potential for a more interactive and dynamic integration between LLMs and KG. Additionally, depending on specific datasets for training limits the broad applicability and flexibility of several studies.

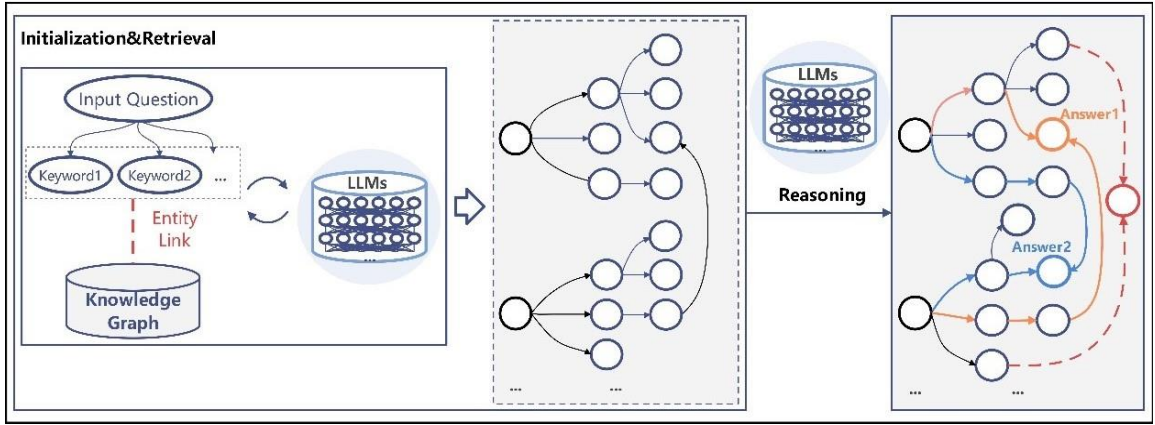

Figure 1: A conceptual demonstration of our scheme, consisting of Initialization, Knowledge Subgraph Retrieval and Reasoning. The blue and orange lines represent the paths and outcomes of the reasoning. The red dashed lines illustrate the augmentation operation based on the implicit knowledge of LLMs.

Building upon these considerations, this study proposes a plug-and-play prompt-based scheme for collaborative reasoning through tight cooperation between KG and LLMs, guiding LLMs to reason by combining external explicit knowledge with their inherent implicit knowledge, as shown in Figure 1. In this scheme, LLMs are engaged to iteratively perform beam search (Jurafsky and Martin, 2009) and proactively explore the graph, consequently retrieving task-specific knowledge subgraphs. Subsequently, it prompts LLMs to combine their inherent implicit knowledge for step-by-step reasoning on the subgraph, concurrently elucidating the reasoning process for clarity. The effectiveness of our scheme is validated on multiple datasets, overall presenting the following advantages: (1) Enhanced Reasoning Ability: Reasoning step by step based on task-targeted knowledge selectively retrieved by LLMs beforehand, our scheme enhances the capability to handle complex tasks. (2) Transparency: Our scheme requires LLMs to clearly delineate their thought process, thereby improving traceability and potentially further sparking the inherent implicit knowledge of LLMs, as shown by the red dashed line in Figure 1. (3) Retrieval and Reasoning Decoupling: By separating knowledge retrieval from reasoning, our scheme enhances both efficiency and flexibility of the reasoning process. (4) High Transferability: Our scheme demonstrates strong adaptability across various KG and LLM combinations without extra training costs.

## 2 Related Work

**Prompt Engineering.** Prompt Engineering focuses on crafting prompts to guide pre-trained models for specific tasks efficiently. Recent strategies such as CoT (Wei et al., 2022) and RAG enhance LLMs' inferencing. CoT improves reasoning by guiding models through a series of step-by-step prompts that align with inference logic, which draw on LLMs' implicit knowledge. However, such knowledge is not always sufficient to satisfy various complex knowledge-intensive tasks. Conversely, RAG boosts inference capabilities by offering pertinent document snippets (Lewis et al., 2020; Mallen et al., 2023; Jiang et al., 2022; Shi et al., 2023; Komeili et al., 2022), which may pose additional challenges due to their lengthy or relevance. Consequently, researchers are currently exploring the application of KG as an alternative.

**KG-Augmented LLMs.** KG stores vast amounts of external world knowledge in structured triplet forms, which have advantages in explicit knowledge representation. Early studies (Zhang et al., 2019; Sun et al., 2021; Wang et al., 2021a; Wang et al., 2021b) concentrate on employing corpora derived from KG triplets for the training of LLMs, consequently enhancing their efficacy in a range of tasks. However, it embeds knowledge from KG directly into LLM parameters, which means any update to the KG requires retraining of the LLMs, posing challenges in terms of flexibility and transparency. Recent studies have integrated related knowledge from KG into prompts to enhance LLM inference, primarily summarized into two strategies: "Reasoning before Retrieval" and "Retrieval before Reasoning".

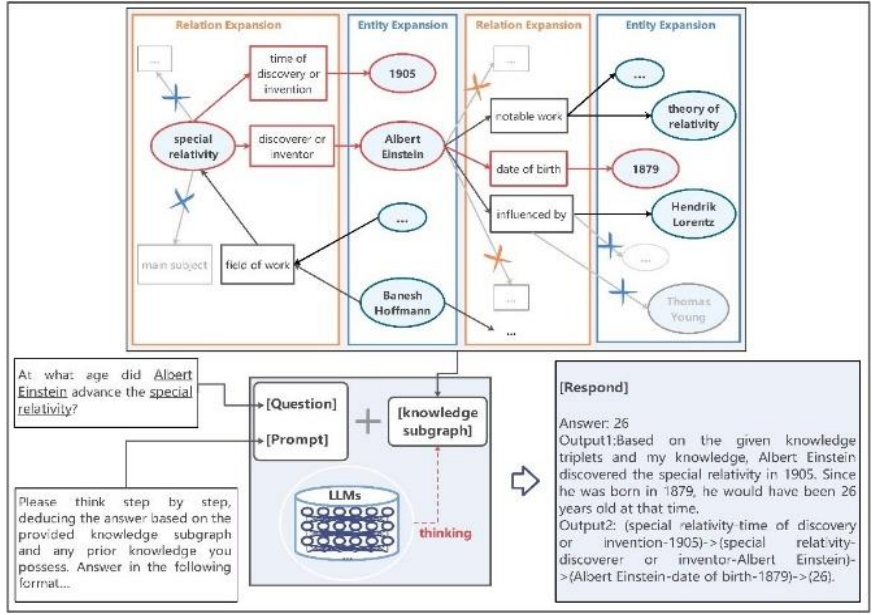

Figure 2 An example of the workflow for the Knowledge Subgraph Retrieval and Reasoning. Entities and relationships depicted in gray are not included in the knowledge subgraph, while those in red represent thought pathways. The response example showcases both the provided answer and the corresponding reasoning path.

In the "Reasoning before Retrieval" strategy, Li et al. (2023a) utilizes LLMs to break down questions into sub-questions and then using fine-tuned models for query generation, retrieving answers for sub-questions to form the final answer. Jiang et al. (2023) starts with the question entity, using LLMs to identify related relationships and entities, extending the search in a single-chain manner until the answer is found. Although this strategy effectively reduces the search range by utilizing LLMs, it may confront scenarios where the answer entity lies outside the pre-established search boundary. Moreover, the methods for directly generating queries as described in the Li et al. (2023a) depend heavily on the completeness of the KG, failing to fully harness the deep and dynamic integration potential between LLMs and KG.

In the "Retrieval before Reasoning" strategy, Baek et al. (2023) samples top K triples for prompting based on semantic similarity, focusing on retrieving single-hop triples which falls short for multi-hop reasoning. Furthermore, it often merely rephrases retrieved facts to respond, not fully leveraging the LLM's profound reasoning abilities. Drawing inspiration from human cognitive behavior, Wang et al. (2023) directs the model to reason step by step based on the retrieved triples. However, its integrated retrieval component incurs additional training costs and exhibits limited transferability across diverse KGs. Compared to these researches, our scheme not only demonstrates substantial flexibility in multi-hop knowledge subgraph retrieval but also effectively activates the LLM's implicit knowledge, facilitating more comprehensive joint reasoning with external explicit knowledge.

## 3 Method

### 3.1 Knowledge-Enhanced LLM Prompting

This study explores the application of KG as a strategy to enhance the reasoning abilities of LLM.

Initially, a KG G is defined as a series of triple sets constructed from the factual information, represented as G={$(e_1 \rightarrow r \rightarrow e_2), |e_1, e_2 \in E, r \in R$}, where E and R respectively signify the sets of entities and relationships. P is then posited as an LLM-based inference system, and Q represents the input question. Traditionally, LLMs primarily depend on internal parameters for reasoning, generating answers A by (1).

$$P_{Reasoning}(A|Q) \quad (1)$$

$$P_{Retrial}(g|Q) \quad (2)$$

$$P_{Reasoning}(A|(p,Q,g)) \quad (3)$$

To mitigate these limitations, this study introduces a prompt-based collaborative reasoning framework, leveraging the synergistic integration of KG and LLMs, as illustrated in Figure 2. Firstly, a retrieval LLM is tasked with extracting a pertinent knowledge subgraph knowledge subgraph g (g $\in$ G) related to question Q and (2) defines the prior distribution over the latent subgraph g conditioned on the question Q. Subsequently, our scheme uses a prompt p to guide the inference LLM to direct the inference LLM in predicting the probability of the answer A considering the specified subgraph g, as illustrated in (3), Specially, our scheme primarily consists of three parts: initialization, knowledge subgraph retrieval, and reasoning.

## 3.2 Initialization

During the initialization phase, our scheme first prompts the underlying LLM to identify and extract pivotal entities from Q. These entities are subsequently linked to the G, represented by a topic entity list $E_Q^0 = \{e_Q^0\}$. The entities within $E_Q^0$ serve as the retrieval starting points on the G, facilitating the construction of the knowledge subgraph.

## 3.3 Knowledge Subgraph Retrieval

Each entity $e_Q^0 \in E_Q^0$ that remains unexpanded is subjected to an iterative expansion of N times, with the primary aim of leveraging the retrieval capabilities of the LLM to search and filter relevant relations and entities. This process encompasses the discovery of information not only directly associated with the input question Q but also indirectly related or implicit knowledge elements. These entities and relationships are subsequently amalgamated into the knowledge subgraph g, laying the foundation for subsequent in-depth understanding and reasoning of question Q.

**Relation expansion** is a beam search process with a width of K guided by retrieval LLM, including both relation searching and relation filtering. In the n-th extended relation search stage of entity $e_Q^{n-1}$, our scheme utilizes predefined relation query template $T_R$ and relation searcher SR to accommodate different KG. SR is employed to search and return all bidirectional relations $R^{C,e_Q^{n-1}}$ connected to the entity $e_Q^{n-1}$, designated as (4) and (5):

$$SR(e,T) = \{r \in R | (e' \rightarrow r \rightarrow e) \vee (e \rightarrow r \rightarrow e')\} \quad (4)$$

$$R^{C,e_Q^{n-1}} = SR(e_Q^{n-1}, T_R) \quad (5)$$

In the n-th stage of extended relation filtering, our scheme employs the retrieval LLM to discern the top K most pertinent relations $R^{K|C,e_Q^{n-1}}$ from $R^{C,e_Q^{n-1}}$, which is based on the given question Q and the current extended entity $e_Q^{n-1}$. The process is articulated in (6), focusing on identifying the relation s that are most conducive to answering the question Q, thereby facilitating the further development of the knowledge subgraph g.

$$P_{Retrial}(R^{K|C,e_Q^{n-1}} | Q, e_Q^{n-1}, R^{C,e_Q^{n-1}}) \quad (6)$$

**Entity expansion** is a beam search process with a width of I guided by retrieval LLM, including both entity searching and entity filtering. Similar to relational search, In the n-th extended entity search stage of entity $e_Q^{n-1}$, our scheme utilizes predefined entity query template $T_R$ and entity searcher SE to accommodate different KG. SE is employed to search and return all entities $E^{C|r,e_Q^{n-1}}$, which is connected to the entity $e_Q^{n-1}$ and relation r, designated as (7) and (8):

$$SE(e,T,r) = \{e' \in E | (e' \rightarrow r \rightarrow e) \vee (e \rightarrow r \rightarrow e')\} \quad (7)$$

$$E^{C|r,e_Q^{n-1}} = SE(e_Q^{n-1}, T_R, r), r \in R^{K|C,e_Q^{n-1}} \quad (8)$$

Consequently, for each entity $e_Q^{n-1}$, the entity set $E^{C,e_Q^{n-1}}$ retrieved and returned by the SE for every relation $r \in R^{K|C,e_Q^{n-1}}$ is represented as follows in (9):

$$E^{C,e_Q^{n-1}} = \bigcup_{r \in R^{K|C,e_Q^{n-1}}} E^{C|r,e_Q^{n-1}} \quad (9)$$

| Method \ Dataset | IO& GPT-3.5-turbo | CoT& GPT-3.5-turbo | SOTA FT | | SOTA Prompt-based | | Ours& GPT-3.5-turbo |
|---|---|---|---|---|---|---|---|
| QALD-10 | 42.0 | 42.9 | SPARQL-QA [1] | 45.4 | - | | 56.2 |
| CWQ | 36.5 | 37.7 | DECAF [2] | 70.4 | - | | 50.4 |
| | | | HGNet [3] | 68.9 | | | |
| | | | NSM [4] | 48.8 | | | |
| | | | PullNet [5] | 45.9 | | | |
| Mintaka | 31.4 | 60.5 | T5 for CBQA [6] | 38.0 | KAPING [11] | 56.8 | 61.7 |
| ZsRE | 27.8 | 28.8 | Single ngram | 74.6 | - | | 87.7 |
| | | | Wikipedia | 74.0 | | | |
| | | | KGI_1 | 72.6 | | | |
| | | | MetaRAG | 71.6 | | | |
| Creak | 89.7 | 90.2 | RACo [7] | 88.2 | - | | 91.5 |
| | | | UNICORN [8] | 79.5 | | | |
| | | | VERA [9] | 82.58 | | | |
| | | | GreaseLM [10] | 77.5 | | | |

Table 1 Experimental results on CWQ, QALD10, Mintaka, ZsRE and Creak. The FT SOTA data of the ZsRE comes from https://paperswithcode.com/sota/slot-filling-on-kilt-zero-shot-re, and the others: [1]: Borroto et al. (2022); [2]: Yu et al. (2023); [3]: Chen et al. (2022); [4]: He et al. (2021);[5]: Sun et al. (2019); [6]: Sen et al. (2022); [7]: Yu et al.(2022); [8]: Lourie et al. (2021); [9]: Liu et al. (2023a); [10]: Zhang et al. (2022); [11]: Baek et al. (2023).

In the n-th extended entity filtering, our scheme employs the retrieval LLM to discern entities $E^{C|r,e_Q^{n-1}}$ that contribute most to answering the question and whose number does not exceed I from $E^{I|(C|r,e_Q^{n-1})}$, which is based on the given question Q, the current extended relation $\in R^{K|C,e_Q^{n-1}}$ and the current extended entity $e_Q^{n-1}$. The process is articulated in (10):

$$P_{Retrial}(E^{I|(C|r,e_Q^{n-1})}|Q,e_Q^{n-1},r,E^{C|r,e_Q^{n-1}}) \quad (10)$$

Consequently, for each entity $e_Q^{n-1}$, the entity set $E^{I|C,e_Q^{n-1}}$ selected for every relation $r \in R^{K|C,e_Q^{n-1}}$ is represented as follows in (11):

$$E^{I|C,e_Q^{n-1}} = \bigcup_{r \in R^{K|C,e_Q^{n-1}}} E^{I|(C|r,e_Q^{n-1})} \quad (11)$$

These chosen entities constitute new triples, as depicted in (12), and are subsequently integrated into the knowledge subgraph g. Concurrently, for entities $E^{I|C,e_Q^{n-1}}$ and $e \notin E_Q^{n-1}$, they are appended to $E_Q^n$ for further expansion of subsequent subgraphs. Note that the number of entities selected within the range [0, I] in practical implementation.

$$\forall r,e(r \in R^{K|C,e_Q^{n-1}} \wedge e \in E^{I|(C|r,e_Q^{n-1})}) \begin{cases} (e_Q^{n-1} \to r \to e) \text{ if } r \text{ points from } e_Q^{n-1} \text{ to } e \\ (e \to r \to e_Q^{n-1}) \text{ if } r \text{ points from } e \text{ to } e_Q^{n-1} \end{cases} \quad (12)$$

### 3.4 Reasoning

Following the described procedure, our scheme successfully acquires the knowledge subgraph g and the probability model (2) can be further refined through a sequence of derivations extending from (6) to (10), culminating in (16). Firstly, Formula (13) defines the set of subsequent entities extended by the entity $e^{n-1}$ and the relation r.

$$X(e^n, r) = \{e | \forall e(e \in E^{I|(C|r,e^{n-1})})\} \quad (13)$$

Then, Formula (14) is employed to calculate the probability model of entity $e^{n-1}$ by conducting single-step relationship expansion and entity expansion for relationship r, focusing on the expansion of g in a single iterative step. Subsequently, Formula (15) delineates the cumulative probability model of the $e^{n-1}$ for the relationship r after n iterations, capturing the comprehensive probabilistic assessment over multiple steps of expansion.

$$T_1(e^{n-1}, r) = P_{Retrial}(R^{K|C,e^{n-1}}|Q,e^{n-1}) * P_{Retrial}(E^{I|(C|r,e^{n-1})}|Q,e^{n-1},r) \quad (14)$$

$$T_2(e^{n-1}, r) = \begin{cases} T_1(e^{n-1},r) * \sum_{r \in R^{K|C,e^{n-1}}} T_2(X(e^n,r),r), & 0 \leq n < N \\ \sum_{r \in R^{K|C,e^{n-1}}} T_1(X(e^n,r),r), & n=N \end{cases} \quad (15)$$

Ultimately, formula (16) summarizes the overall probability model of the entire knowledge subgraph g under the conditions of a given question

| Input Question | 1. How many years did the second oldest dog in the world live? | 2. What is the ruling party of the government now in South Korea? |
|---|---|---|
| IO& GPT-3.5-turbo | I'm sorry, but I don't have access to specific information about the age of the second-oldest dog in the world. | The ruling party of the government in South Korea is the Democratic Party of Korea. |
| CoT& GPT-3.5-turbo | First, we need to find out the age of the second oldest dog in the world. Second, we need to determine how many years the dog lived. The answer is the age of the second oldest dog in the world. | First, the ruling party of the government in South Korea is the Democratic Party of Korea. The answer is the Democratic Party of Korea. |
| Ours& GPT-3.5-turbo | Answer:29 | Answer: People Power Party |
|  | Output 1: Based on the given knowledge triplets and my knowledge, the second oldest dog in the world is Bluey. Bluey was born on June 7, 1910, and died on November 14, 1939. Therefore, Bluey lived for 29 years. | Output 1: Based on the given knowledge triplets and my knowledge, Yoon Suk Yeol is the head of government in South Korea and a member of the People Power Party. Therefore, the ruling party of the government now in South Korea is the People Power Party. |
|  | Output 2:(Bluey-date of birth-1910-06-07T00:00:00Z)->(Bluey-date of death-1939-11-14T00:00:00Z)->(Bluey-lived for 29 years). | Output 2: (South Korea-head of government-Yoon Suk Yeol)->(Yoon Suk Yeol-member of political party-People Power Party)->(People Power Party). |

Table 2 Case examples of outputs by different methods. The red color highlights the incorrect answer, and the blue color highlights the correct answer.

Q, taking into account all potential paths of relationship and entity expansion.

$$P_{Retrial}(g|Q) = \sum_{e_Q^0 \in E_Q^0} \sum_{r \in R^{k|C, e_Q^0}} T_2(e_Q^0, r) \quad (16)$$

Following the successful establishment of the knowledge subgraph g, our scheme directs the reasoning LLM to comprehend g using succinct triplets and prompts the LLM to explicitly describe its reasoning path while thinking step by step. It encourages the LLM to integrate its inherent implicit knowledge for collaborative reasoning, potentially enabling it to identify new links between nodes not present in the existing subgraph or to expand new nodes from existing nodes. Such a strategy is instrumental in producing more accurate and insightful responses, effectively harnessing the full reasoning power of the LLM. In culmination, the overall probabilistic model for the reasoned answer A is encapsulated in Equation (17).

$$P = P_{Reasoning}(A|(p,Q,g)) * P_{Retrial}(g|Q)$$
$$= P_{Reasoning}(A|(p,Q,g)) * \sum_{e_Q^0 \in E_Q^0} \sum_{r \in R^{k|C, e_Q^0}} T_2(e_Q^0, r) \quad (17)$$

## 4 Experiments

### 4.1 Setup

**Dataset.** This study evaluates the proposed scheme across multiple datasets, involving 3 KBQA datasets, 1 slot filling dataset, 1 fact-checking dataset, and 1 medical QA dataset: CWQ (Talmor and Berant, 2018), QALD10 (Perevalov et al., 2022), Mintaka (Sen et al., 2022), ZsRE (Petroni et al., 2021), Creak (Onoe et al., 2021) and ExplainCPE (Li et al., 2023b). Specifically, 1000 questions from CWQ, Mintaka and 400 from ExplainCPE, are randomly selected for testing. In addition, this study established SMKG, constructed using vertical pharmaceutical websites as its data sources. SMKG is used to support reasoning on the ExplainCPE dataset and is stored in Neo4j.

**Details.** In this study, standard input-output (IO) prompts (Brown et al., 2020) and chain-of-thought (CoT) prompts (Wei et al., 2022) were used as baseline methods, and use exact match accuracy (Hits@1) as the evaluation metric. For datasets CWQ, QALD-10, Mintaka, ZsRE and Creak, comparisons were made with both baseline methods and previous fine-tuning state-of-the-art (SOTA) works. The expansion depth of relations was set to 2 in the ExplainCPE and Mintaka experiments, while the default setting for other experiments was 3. The expansion width of entities was set to 10 across all experiments, and the number of expansion iterations was set to 2. The above settings are the default configurations for the experiments unless specified otherwise. To ensure diversity in generation and reproducible results, the temperature parameter was set to 0.4 for the expansion process and 0 for the reasoning process.

Additionally, considering that ExplainCPE is not constructed upon SMKG, the keywords extracted by LLM might not fully align with entities in SMKG. Consequently, the Initialization of the experiment on ExplainCPE employed a BERT encoder. It transformed the extracted keywords and entities in the KG into word vectors, followed by entity linking by comparing their cosine similarities.

### 4.2 Results

**Compared to baseline methods without external knowledge.** As depicted in Table 1, our scheme surpasses all baseline methods across all datasets. In particular, it shows a significant increase in performance in the ZsRE dataset, with 60% over the IO method and 58.9% over the CoT method. This phenomenon indicates that integrating knowledge from external KG substantially enhances the capabilities of LLMs in performing a variety of knowledge-intensive tasks.

**Compared with previous SOTA.** Despite the inherent advantages of fine-tuning methods trained with evaluation data, our scheme nevertheless outperformed advanced fine-tuning methods on the Creak and QALD-10 datasets, which demonstrates the applicability and effectiveness of LLMs in actively participating in reasoning, as shown in Table 1. In particular, the performance on the QALD-10 dataset is improved by more than 10%. Furthermore, compared to the prompt-based SOTA, our scheme demonstrates improved performance on Mintaka, highlighting its advantages. However, due to experimental constraints, the implementation of our scheme in the CWQ leverages Wikidata as the underlying KG. Although it did not attain the same level of performance as several advanced SOTA reliant on Freebase, this gap is acceptable and unavoidable, considering that the CWQ dataset is built on Freebase. This observation further also indicates that the matching degree and quality of the knowledge graph may affect the performance of our scheme in complex reasoning tasks.

**Case Study.** As shown in Table 2, two examples are selected for analysis. The first example illustrates that both the IO and CoT methods struggle to answer questions without sufficient background knowledge. Through collaboration with a knowledge graph, our scheme empowers the retrieval LLM to effectively retrieve crucial information, such as the birth and death dates of the second oldest dog, resulting in accurate responses. In the second example, both the Input-Output (IO) and CoT methods produce inaccurate answers. Specifically, despite the absence of a direct knowledge triple link to the question's answer, our scheme effectively leverages implicit knowledge to correctly identify ' People Power Party ' as the answer. GPT-3.5 in the baseline method erroneously identified the former ruling party as the 'Democratic Party of Korea' due to the knowledge not being updated promptly.

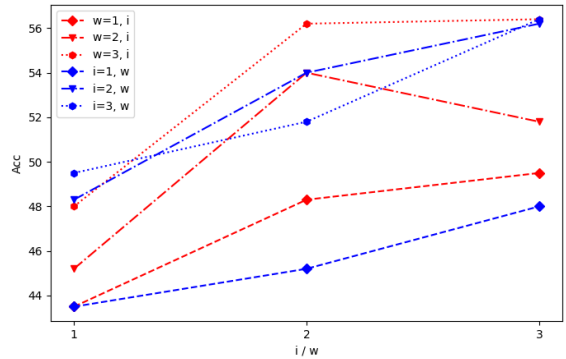

Figure 3 Performance on different search Iteration numbers and expansion width.

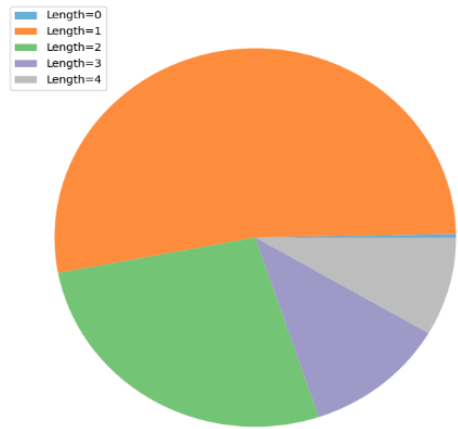

Figure 4 The statistics on the relation numbers of SPARQL queries within QALD-10.

**Number of Iterations and expansion width.** Further exploration and analysis were conducted on the QALD-10 dataset. Experiments were carried out under settings where the number of expansion iterations and the width of relational expansion ranged from 1 to 3, as shown in Figure 3. The results explicate an overall augmentation in the efficacy of our proposed scheme with increasing iterations and widths. Furthermore, they also manifest a positive correlation between the increase in the width and the heightened probability of the knowledge subgraph encompassing pivotal information requisite for answering the queries. However, a deceleration in performance enhancement is discernible when the number of iterations surpasses two. This observed phenomenon can plausibly be attributed to the inherent characteristics of the questions in QALD-10, which predominantly necessitate an inference length not exceeding two, as elucidated in Figure 4.
**Flexibility and Practicality.** Further experiments about different KG and LLMs were conducted on ExplainCPE and QALD-10, as shown in Figure 5 and Figure 6. In the ExplainCPE experiment, our scheme demonstrated an improvement in accuracy of over 8% compared to the baseline method, under identical conditions using GPT3.5, thereby showcasing its adaptability with various KGs. Moreover, it is worth noting that the performance of this scheme on ExplainCPE was lower than that of GPT-4, which may be due to the extensive corpus used in GPT-4's pre-training, covering the knowledge required to answer questions in that dataset. In the QALD-10 experiment, we utilized both GPT-3.5 and GPT-4 as distinct backbone models. The experimental results demonstrate that our scheme successfully enhances the reasoning capabilities of these two different LLMs, and also exhibits portability across these different models. In addition, the performance achieved with our scheme on GPT-3.5 surpasses that of the CoT method using GPT-4. This outcome suggests that our scheme has the potential to elevate the effectiveness of smaller LLMs, enabling them to compete more effectively with larger LLMs.

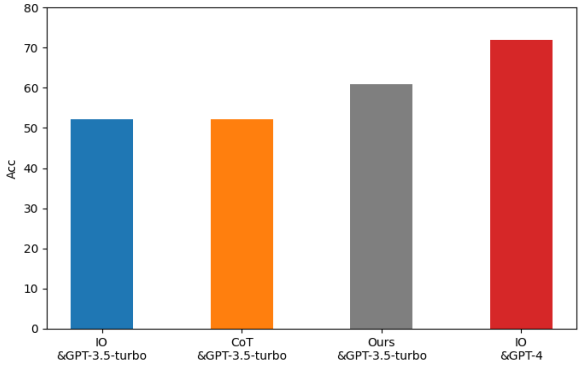

Figure 5 Experimental results on ExplainCPE.

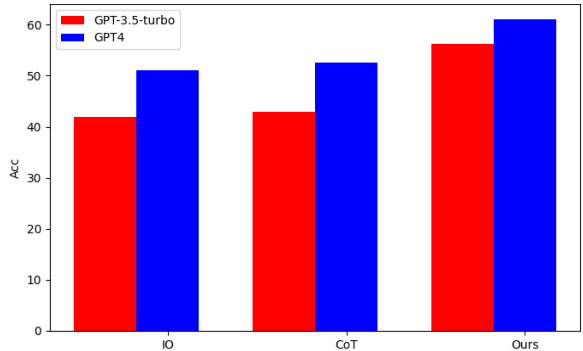

Figure 6 Performance with different LLM on QALD-10

## 5 Conclusions

This study proposes a versatile training-free reasoning scheme that achieves tight collaboration between KG and LLMs, as well as the decoupling of reasoning and retrieval. The central principle of this scheme is its dual capability to enhance the reasoning abilities of LLMs and to effectively excavate and utilize the latent knowledge within LLMs. Through this deep collaborative method, our scheme achieves more accurate and reliable knowledge reasoning. The effectiveness of our scheme is demonstrated through its superior performance across multiple datasets, showcasing its ability to handle specialized and complex tasks more efficiently.

## References

Tom Brown, Benjamin Mann, Nick Ryder, Melanie Subbiah, Jared D Kaplan, Prafulla Dhariwal, Arvind Neelakantan, Pranav Shyam, Girish Sastry, Amanda Askell, Sandhini Agarwal, Ariel Herbert-Voss, Gretchen Krueger, Tom Henighan, Rewon Child, Aditya Ramesh, Daniel Ziegler, Jeffrey Wu, Clemens Winter, Chris Hesse, Mark Chen, Eric Sigler, Mateusz Litwin, Scott Gray, Benjamin Chess, Jack Clark, Christopher Berner, Sam McCandlish, Alec Radford, Ilya Sutskever, and Dario Amodei. 2020. Language models are few-shot learners. Advances in Neural Information Processing Systems, 33:1877–1901.

Manuel Borroto, Francesco Ricca, Bernardo Cuteri, and Vito Barbara. 2022. SPARQLQA enters the QALD challenge. In Proceedings of the 7th Natural Language Interfaces for the Web of Data (NLIWoD) co-located with the 19th European Semantic Web Conference, Hersonissos, Greece, volume 3196 of CEUR Workshop Proceedings, pp. 25–31.

Jinheon Baek, Alham Fikri Aji, and Amir Saffari. 2023. Knowledge-augmented language model prompting

for zero-shot knowledge graph question answering. In Proceedings of the 1st Workshop on Natural Language Reasoning and Structured Explanations (NLRSE), pages 78–106, Toronto, Canada.

Yongrui Chen, Huiying Li, Guilin Qi, Tianxing Wu, and Tenggou Wang. 2022. Outlining and Filling: Hierarchical Query Graph Generation for Answering Complex Questions Over Knowledge Graphs[J]. IEEE Transactions on Knowledge and Data Engineering, vol. 35, no. 8, pp. 8343-8357.

Aakanksha Chowdhery, Sharan Narang, Jacob Devlin, Maarten Bosma, Gaurav Mishra, Adam Roberts, Paul Barham, Hyung Won Chung, Charles Sutton, Sebastian Gehrmann, Parker Schuh, Kensen Shi, Sasha Tsvyashchenko, Joshua Maynez, Abhishek Rao, Parker Barnes, Yi Tay, Noam Shazeer, Vinodkumar Prabhakaran, Emily Reif, Nan Du, Ben Hutchinson, Reiner Pope, James Bradbury, Jacob Austin, Michael Isard, Guy Gur-Ari, Pengcheng Yin, Toju Duke, Anselm Levskaya, Sanjay Ghemawat, Sunipa Dev, Henryk Michalewski, Xavier Garcia, Vedant Misra, Kevin Robinson, Liam Fedus, Denny Zhou, Daphne Ippolito, David Luan, Hyeontaek Lim, Barret Zoph, Alexander Spiridonov, Ryan Sepassi, David Dohan, Shivani Agrawal, Mark Omernick, Andrew M. Dai, Thanumalayan Sankaranarayana Pillai, Marie Pellat, Aitor Lewkowycz, Erica Moreira, Rewon Child, Oleksandr Polozov, Katherine Lee, Zongwei Zhou, Xuezhi Wang, Brennan Saeta, Mark Diaz, Orhan Firat, Michele Catasta, Jason Wei, Kathy Meier-Hellstern, Douglas Eck, Jeff Dean, Slav Petrov, and Noah Fiedel. 2023. Palm: Scaling language modeling with pathways. Journal of Machine Learning Research, 24(240), 1-113.

Marina Danilevsky, Kun Qian, Ranit Aharonov, Yannis Katsis, Ban Kawas, and Prithviraj Sen. 2020. A Survey of the State of Explainable AI for Natural Language Processing. In Proceedings of the 1st Conference of the Asia-Pacific Chapter of the Association for Computational Linguistics and the 10th International Joint Conference on Natural Language Processing, 447–459.

Zhengxiao Du, Yujie Qian, Xiao Liu, Ming Ding, Jiezhong Qiu, Zhilin Yang, and Jie Tang. 2022. GLM: General Language Model Pretraining with Autoregressive Blank Infilling. In Proceedings of the 60th Annual Meeting of the Association for Computational Linguistics (Volume 1: Long Papers), pages 320–335, Dublin, Ireland. Association for Computational Linguistics.

Gaole He, Yunshi Lan, Jing Jiang, Wayne Xin Zhao, and Ji-Rong Wen. 2021. Improving Multi-hop Knowledge Base Question Answering by Learning Intermediate Supervision Signals. In Proceedings of the 14th ACM International Conference on Web Search and Data Mining (WSDM '21). Association for Computing Machinery, New York, NY, USA, 553–561.

Dan Jurafsky and James H. Martin. 2009. Speech and language processing: an introduction to natural language processing, computational linguistics, and speech recognition, 2nd Edition. Prentice Hall series in artificial intelligence. Prentice Hall, Pearson Education International.

Zhengbao Jiang, Luyu Gao, Zhiruo Wang, Jun Araki, Haibo Ding, Jamie Callan, and Graham Neubig. 2022. Retrieval as Attention: End-to-end Learning of Retrieval and Reading within a Single Transformer. In Proceedings of the 2022 Conference on Empirical Methods in Natural Language Processing, pages 2336–2349, Abu Dhabi, United Arab Emirates.

Jinhao Jiang, Kun Zhou, Zican Dong, Keming Ye, Xin Zhao, and Ji-Rong Wen. 2023. StructGPT: A General Framework for Large Language Model to Reason over Structured Data. In Proceedings of the 2023 Conference on Empirical Methods in Natural Language Processing, pages 9237–9251, Singapore.

Ziwei Ji, Nayeon Lee, Rita Frieske, Tiezheng Yu, Dan Su, Yan Xu, Etsuko Ishii, Ye Jin Bang, Andrea Madotto, and Pascale Fung. 2023. Survey of Hallucination in Natural Language Generation. ACM Computing Surveys, 55(12): 1–38.

Mojtaba Komeili, Kurt Shuster, and Jason Weston. 2022. Internet-Augmented Dialogue Generation. In Proceedings of the 60th Annual Meeting of the Association for Computational Linguistics (Volume 1: Long Papers), pages 8460–8478, Dublin, Ireland. Association for Computational Linguistics.

Jean Kaddour, Joshua Harris, Maximilian Mozes, Herbie Bradley, and Roberta Raileanu, Robert McHardy. 2023. Challenges and Applications of Large Language Models. ArXiv, abs/2307.10169.

Anis Koubaa, Wadii Boulila, Lahouari Ghouti, Ayyub Alzahem, and Shahid Latif. 2023. Exploring ChatGPT Capabilities and Limitations: A Survey. IEEE Access, vol. 11, pp. 118698-118721.

Patrick Lewis, Ethan Perez, Aleksandra Piktus, Fabio Petroni, Vladimir Karpukhin, Naman Goyal, Heinrich Küttler, Mike Lewis, Wen-tau Yih, Tim Rocktäschel, Sebastian Riedel, and Douwe Kiela. 2020. Retrieval-augmented generation for knowledge-intensive nlp tasks. Advances in Neural Information Processing Systems, 33: 9459–9474.

Nicholas Lourie, Ronan Le Bras, Chandra Bhagavatula, and Yejin Choi. 2021. Unicorn on rainbow: A universal commonsense reasoning model on a new multitask benchmark. In Proceedings of the AAAI Conference on Artificial Intelligence (Vol. 35, No. 15, pp. 13480-13488).

Xingxuan Li, Ruochen Zhao, Yew Ken Chia, Bosheng Ding, Lidong Bing, Shafiq Joty, and Soujanya Poria. 2023a. Chain-of-Knowledge: Grounding Large Language Models via Dynamic Knowledge Adapting over Heterogeneous Sources. arXiv preprint arXiv:2305.13269.

Dongfang Li, Jindi Yu, Baotian Hu, Zhenran Xu, and Min Zhang. 2023b. ExplainCPE: A Free-text Explanation Benchmark of Chinese Pharmacist Examination. In Findings of the Association for Computational Linguistics: EMNLP 2023, pages 1922–1940, Singapore. Association for Computational Linguistics.

Jiacheng Liu, Wenya Wang, Dianzhuo Wang, Noah Smith, Yejin Choi, and Hannaneh Hajishirzi. 2023a. Vera: A General-Purpose Plausibility Estimation Model for Commonsense Statements. In Proceedings of the 2023 Conference on Empirical Methods in Natural Language Processing, pages 1264–1287, Singapore. Association for Computational Linguistics.

Nelson F. Liu, Kevin Lin, John Hewitt, Ashwin Paranjape, Michele Bevilacqua, Fabio Petroni, and Percy Liang. 2023b. Lost in the Middle: How Language Models Use Long Contexts. arXiv preprint arXiv:2307.03172.

Alex Mallen, Akari Asai, Victor Zhong, Rajarshi Das, Daniel Khashabi, and Hannaneh Hajishirzi. 2023. When Not to Trust Language Models: Investigating Effectiveness of Parametric and Non-Parametric Memories. In Proceedings of the 61st Annual Meeting of the Association for Computational Linguistics (Volume 1: Long Papers), pages 9802–9822, Toronto, Canada. Association for Computational Linguistics.

Yasumasa Onoe, Michael J. Q. Zhang, Eunsol Choi, and Greg Durrett. 2021. CREAK: A dataset for commonsense reasoning over entity knowledge. Proceedings of the Neural Information Processing Systems Track on Datasets and Benchmarks: Dataset and Benchmark Poster Session 1.

Long Ouyang, Jeffrey Wu, Xu Jiang, Diogo Almeida, Carroll Wainwright, Pamela Mishkin, Chong Zhang, Sandhini Agarwal, Katarina Slama, Alex Ray, John Schulman, Jacob Hilton, Fraser Kelton, Luke Miller, Maddie Simens, Amanda Askell, Peter Welinder, Paul F. Christiano, Jan Leike, and Ryan Lowe. 2022. Training language models to follow instructions with human feedback. Advances in Neural Information Processing Systems, 35, 27730-27744.

OpenAI. 2023. GPT-4 Technical Report. ArXiv, abs/2303.08774.

Fabio Petroni, Aleksandra Piktus, Angela Fan, Patrick Lewis, Majid Yazdani, Nicola De Cao, James Thorne, Yacine Jernite, Vladimir Karpukhin, Jean Maillard, Vassilis Plachouras, Tim Rocktäschel, and Sebastian Riedel. 2021. KILT: a Benchmark for Knowledge Intensive Language Tasks. In Proceedings of the 2021 Conference of the North American Chapter of the Association for Computational Linguistics: Human Language Technologies, pages 2523–2544, Online. Association for Computational Linguistics.

Aleksandr Perevalov, Dennis Diefenbach, Ricardo Usbeck, and Andreas Both. Qald-9-plus: A multilingual dataset for question answering over dbpedia and wikidata translated by native speakers. 2022. IEEE 16th International Conference on Semantic Computing (ICSC), Laguna Hills, CA, USA, 2022, pp. 229-234.

Shirui Pan, Linhao Luo, Yufei Wang, Chen Chen, Jiapu Wang, and Xindong Wu. 2023. Unifying large language models and knowledge graphs: A roadmap. arXiv preprint arXiv:2306.08302.

Haitian Sun, Tania Bedrax-Weiss, and William Cohen. 2019. PullNet: Open Domain Question Answering with Iterative Retrieval on Knowledge Bases and Text. In Proceedings of the 2019 Conference on Empirical Methods in Natural Language Processing and the 9th International Joint Conference on Natural Language Processing (EMNLP-IJCNLP), pages 2380–2390, Hong Kong, China. Association for Computational Linguistics.

Yu Sun, Shuohuan Wang, Shikun Feng, Siyu Ding, Chao Pang, Junyuan Shang, Jiaxiang Liu, Xuyi Chen, Yanbin Zhao, Yuxiang Lu, Weixin Liu, Zhihua Wu, Weibao Gong, Jianzhong Liang, Zhizhou Shang, Peng Sun, Wei Liu, Xuan Ouyang, Dianhai Yu, Hao Tian, Hua Wu, and Haifeng Wang. 2021. ERNIE 3.0: Large-scale knowledge enhanced pre-training for language understanding and generation. arXiv preprint arXiv:2107.02137

Priyanka Sen, Alham Fikri Aji, and Amir Saffari. 2022. Mintaka: A Complex, Natural, and Multilingual Dataset for End-to-End Question Answering. In Proceedings of the 29th International Conference on Computational Linguistics, pages 1604–1619, Gyeongju, Republic of Korea. International Committee on Computational Linguistics.

Weijia Shi, Sewon Min, Michihiro Yasunaga, Minjoon Seo, Rich James, Mike Lewis, Luke Zettlemoyer, and Wen-tau Yih. 2023. Replug: Retrieval-augmented black-box language models. arXiv preprint arXiv:2301.12652.

Alon Talmor and Jonathan Berant. 2018. The Web as a Knowledge-Base for Answering Complex Questions. In Proceedings of the 2018 Conference of the North American Chapter of the Association for Computational Linguistics: Human Language Technologies, Volume 1 (Long Papers), pages 641–651, New Orleans, Louisiana. Association for Computational Linguistics.

Romal Thoppilan, Daniel De Freitas, Jamie Hall, Noam Shazeer, Apoorv Kulshreshtha, Heng-Tze Cheng, Alicia Jin, Taylor Bos, Leslie Baker, Yu Du, YaGuang Li, Hongrae Lee, Huaixiu Steven Zheng, Amin Ghafouri, Marcelo Menegali, Yanping Huang, Maxim Krikun, Dmitry Lepikhin, James Qin, Dehao Chen, Yuanzhong Xu, Zhifeng Chen, Adam Roberts, Maarten Bosma, Vincent Zhao, Yanqi Zhou, Chung-Ching Chang, Igor Krivokon, Will Rusch, Marc Pickett, Pranesh Srinivasan, Laichee Man, Kathleen Meier-Hellstern, Meredith Ringel Morris, Tulsee Doshi, Renelito Delos Santos, Toju Duke, Johnny Soraker, Ben Zevenbergen, Vinodkumar Prabhakaran, Mark Diaz, Ben Hutchinson, Kristen Olson, Alejandra Molina, Erin Hoffman-John, Josh Lee, Lora Aroyo, Ravi Rajakumar, Alena Butryna, Matthew Lamm, Viktoriya Kuzmina, Joe Fenton, Aaron Cohen, Rachel Bernstein, Ray Kurzweil, Blaise Aguera-Arcas, Claire Cui, Marian Croak, Ed Chi, and Quoc Le. 2022. LaMDA: Language Models for Dialog Applications. arXiv preprint arXiv:2201.08239.

Hugo Touvron, Thibaut Lavril, Gautier Izacard, Xavier Martinet, Marie-Anne Lachaux, Timothée Lacroix, Baptiste Rozière, Naman Goyal, Eric Hambro, Faisal Azhar, Aurelien Rodriguez, Armand Joulin, Edouard Grave, and Guillaume Lample. 2023. LLaMA: Open and Efficient Foundation Language Models. arXiv preprint arXiv:2302.13971.

Ruize Wang, Duyu Tang, Nan Duan, Zhongyu Wei, Xuanjing Huang, Jianshu Ji, Guihong Cao, Daxin Jiang, and Ming Zhou. 2021a. K-Adapter: Infusing Knowledge into Pre-Trained Models with Adapters. In Findings of the Association for Computational Linguistics: ACL-IJCNLP 2021, pp. 1405–1418.

Xiaozhi Wang, Tianyu Gao, Zhaocheng Zhu, Zhengyan Zhang, Zhiyuan Liu, Juanzi Li, and Jian Tang. 2021b. Kepler: A unified model for knowledge embedding and pre-trained language representation. Transactions of the Association for Computational Linguistics, 9:176–194.

Jason Wei, Xuezhi Wang, Dale Schuurmans, Maarten Bosma, Brian Ichter, Fei Xia, Ed Chi, Quoc Le, and Denny Zhou. 2022. Chain-of-thought prompting elicits reasoning in large language models. Advances in Neural Information Processing Systems, 35, 24824-24837.

Jianing Wang, Qiushi Sun, Nuo Chen, Xiang Li, and Ming Gao. 2023. Boosting Language Models Reasoning with Chain-of-Knowledge Prompting. arXiv preprint arXiv:2306.06427.

Tianbao Xie, Chen Henry Wu, Peng Shi, Ruiqi Zhong, Torsten Scholak, Michihiro Yasunaga, ChienSheng Wu, Ming Zhong, Pengcheng Yin, Sida I. Wang, Victor Zhong, Bailin Wang, Chengzu Li, Connor Boyle, Ansong Ni, Ziyu Yao, Dragomir Radev, Caiming Xiong, Lingpeng Kong, Rui Zhang, Noah A. Smith, Luke Zettlemoyer, and Tao Yu. 2022. UnifiedSKG: Unifying and multitasking structured knowledge grounding with text-to-text language models. In Proceedings of the 2022 Conference on Empirical Methods in Natural Language Processing, pp. 602–631, Abu Dhabi, United Arab Emirates.

Linyao Yang, Hongyang Chen, Zhao Li, Xiao Ding, and Xindong Wu. 2023. ChatGPT is not Enough: Enhancing Large Language Models with Knowledge Graphs for Fact-aware Language Modeling. arXiv preprint arXiv:2306.11489.

Wenhao Yu, Chenguang Zhu, Zhihan Zhang, Shuohang Wang, Zhuosheng Zhang, Yuwei Fang, and Meng Jiang. 2022. Retrieval augmentation for commonsense reasoning: A unified approach. In Proceedings of the 2022 Conference on Empirical Methods in Natural Language Processing, pp.4364–4377, Abu Dhabi, United Arab Emirates.

Donghan Yu, Sheng Zhang, Patrick Ng, Henghui Zhu, Alexander Hanbo Li, Jun Wang, Yiqun Hu, William Wang, Zhiguo Wang, and Bing Xiang. 2023. Decaf: Joint decoding of answers and logical forms for question answering over knowledge bases, in ICLR 2023.

Zhengyan Zhang, Xu Han, Zhiyuan Liu, Xin Jiang, Maosong Sun, and Qun Liu. 2019. ERNIE: Enhanced Language Representation with Informative Entities. In Proceedings of the Annual Meeting of the Association for Computational Linguistics, 1441–1451.

Xikun Zhang, Antoine Bosselut, Michihiro Yasunaga, Hongyu Ren, Percy Liang, Christopher D Manning, and Jure Leskovec. 2022. GreaseLM: Graph REASoning enhanced language models. In International Conference on Learning Representations.